\def\year{2024}\relax
\documentclass[letterpaper]{article} 
\usepackage{aaai22}  
\usepackage{times}  
\usepackage{helvet}  
\usepackage{courier}  
\usepackage[hyphens]{url}  
\usepackage{graphicx} 
\usepackage{natbib}  
\usepackage{caption} 

\usepackage{booktabs} 
\usepackage{amsmath, bm}
\usepackage{amsfonts}
\usepackage{amsthm}
\usepackage{multirow}
\usepackage{amssymb}
\usepackage{dsfont}
\usepackage{subcaption}
\usepackage{xcolor}
\usepackage{xspace}

\DeclareCaptionStyle{ruled}{labelfont=normalfont,labelsep=colon,strut=off} 
\usepackage[ruled, vlined]{algorithm2e}

\def \Lt {\mathcal{L}_{\text{T}}}
\def \Lv {\mathcal{L}_{\text{V}}}

\def \Ours {CHAIN\xspace}
\def \Tin {T_1}
\def \Tout {T_2}
\def \eg {\textit{e.g.}}
\def \ie {\textit{i.e.}}

\DeclareMathOperator*{\argmin}{arg\,min}
\definecolor{grey}{HTML}{A9A9A9}

\newcommand\best[1]{\textcolor{red}{#1}}
\newcommand\secbest[1]{\textcolor{blue}{#1}}
\newcommand\normal[1]{\textcolor{grey}{#1}}
\title{Taming Small-sample Bias in Low-budget Active Learning}
\author{
    Linxin Song \textsuperscript{\rm 1} \ 
    Jieyu Zhang \textsuperscript{\rm 2} \ 
    Xiaotian Lu \textsuperscript{\rm 3} \ 
    Tianyi Zhou \textsuperscript{\rm 4}
}
\affiliations{
    \textsuperscript{\rm 1} Waseda University \ 
    \textsuperscript{\rm 2} University of Washington \ 
    \textsuperscript{\rm 3} Kyoto University \ 
    \textsuperscript{\rm 4} University of Maryland, College Park \ 

    songlx.imse.gt@ruri.waseda.jp, jieyuz2@cs.washington.edu, lu@ml.ist.i.kyoto-u.ac.jp, tianyi.david.zhou@gmail.com
}

\usepackage{bibentry}

\begin{document}

\maketitle

\begin{abstract}
Active learning (AL) aims to minimize the annotation cost by only querying a few informative examples for each model training stage. 
However, training a model on a few queried examples suffers from the small-sample bias.
In this paper, we address this small-sample bias issue in low-budget AL by exploring a regularizer called Firth bias reduction, which can provably reduce the bias during the model training process but might hinder learning if its coefficient is not adaptive to the learning progress. 
Instead of tuning the coefficient for each query round, which is sensitive and time-consuming, we propose the curriculum Firth bias reduction (CHAIN) that can automatically adjust the coefficient to be adaptive to the training process.
Under both deep learning and linear model settings, experiments on three benchmark datasets with several widely used query strategies and hyperparameter searching methods show that CHAIN can be used to build more efficient AL and can substantially improve the progress made by each active learning query. 
\end{abstract}

\section{Introduction}
The last decade has witnessed the emergence of deep learning as the dominant force in advancing machine learning and its applications.
To reduce the data annotation cost, researchers and practitioners have resorted to active learning (AL), which aims to reduce the total number of annotations by iteratively querying the most informative annotations for the learning process from an oracle~\citep{settles2009active, schroder2020survey}.
AL uses a finite budget to select and label a subset of a data pool for downstream supervised learning.
By iteratively training a model on a current set of labeled data and querying labels for more new points, this paradigm yields models that use as fewer data as possible to achieve nearly the same accuracy as classifiers with unlimited labeling budgets.

Many traditional active learning approaches are based on uncertainty sampling (e.g., ~\citet{lewis1995sequential, TobiasScheffer2001ActiveHM, AronCulotta2005ReducingLE, AjayJoshi2009MulticlassAL, XinLi2013AdaptiveAL, DanielGissin2019DiscriminativeAL, SamarthSinha2019VariationalAA}), wherein the learner queries the most uncertain examples for annotations, presumably because such examples are most informative to the learner. 
However, such a type of approach primarily relies on the quality of estimated uncertainty.
Besides, existing deep active learning strategies require a large amount of labeled examples to work properly~\citep{MichelleYuan2020ColdstartAL, KossarPourahmadi2021ASB}.
However, the labeling budget might be limited in practice.
Under this low-budget regime, it is well-established that the conventional Maximum Likelihood Estimation (MLE) used for training the model is biased with a term of $O(N^{-1})$~\citep{DavidCox1968AGD, MJBox1971BiasIN, JohnWhitehead1986OnTB, EwoutWSteyerberg1999StepwiseSI}.
Such \textit{small-sample bias} in low-budget setup might deteriorate the performance of uncertainty-based AL due to (1) an inaccurate model trained with MLE; and (2) a poor uncertainty estimation caused by the inaccurate model. 

To tackle the small-sample bias in AL, we, for the first time, leverage the Firth bias reduction~\citep{firth1993bias, ghaffari2021importance} as a plug-in regularizer for the MLE-based model training.
It reduces the small-sample bias by applying a regularizer in addition to the cross-entropy loss, which computes the log-determinant of Fisher information matrix $\log(\text{det}(F))$.
However, the performance of the Firth regularizer is sensitive to its coefficient~\citep{ghaffari2021importance}, a hyperparameter that controls the strength of the regularizer.
Thus, how to search for the optimal coefficient becomes crucial when adopting the Firth regularizer.
Grid search, as a standard solution for hyperparameter search, requires repeating the training process multiple times to search for the best coefficient. It is even more time-consuming in AL since the model has to be re-trained after each time the new queried data is added, and so does the hyperparameter search process. 

To reduce the time complexity, we propose to optimize the Firth coefficient on the fly without heavy parameter tuning.
Specifically, we formulate the optimization of both the model parameters and the Firth coefficient as a bilevel optimization where the inner level optimization corresponds to the model parameters, and the outer one optimizes the Firth coefficient.
Such a bilevel optimization can be solved efficiently by interleaving the gradient descent steps of the model parameter and the Firth coefficient. 
In experiments, it only requires a few gradient descent steps (\eg, 50) to obtain a good coefficient.

Furthermore, the impact of the bias can vary drastically across different stages of the training process, while over-regularization might degrade the generalization performance, and under-regularization might not be able to alleviate the small-sample bias issue effectively. 
Hence, the strength of Firth regularization (\ie, the Firth coefficient) should be adaptively adjusted rather than fixed.
To this end, we propose a simple curriculum of the Firth coefficient that dynamically adjust the Firth coefficient to be adaptive to the model's training process. 

Finally, we incorporate the aforementioned components seamlessly in CHAIN (\textbf{C}urriculum firt\textbf{H} bi\textbf{A}s Reduct\textbf{I}o\textbf{N}), which is a general query-strategy-agnostic method for tackling the small-sample bias issue in low-budget AL effectively and efficiently.
We conduct extensive experiments on three benchmark datasets with five widely used query strategies under deep learning and linear model settings.
Compared with the baselines, CHAIN achieves the highest performance in most of the query rounds.
To help understand the curriculum of the Firth coefficient, we provide the analysis of the coefficient curriculum on both deep model and logistic regression, showing that the curriculum exists during training and is related to the size of the queried training set.
We also conduct ablation studies for major hyperparameters used in CHAIN to demonstrate the consistent superiority of CHAIN under different hyperparameter choices.
Our main contributions are summarized as: (1) we propose a query-strategy-agnostic framework that tackles the small-sample bias issue in low-budget AL via a plug-in regularization and efficient optimization of the coefficient of the regularizer, (2) we further improve the framework by a simple curriculum of the coefficient that adaptively changes the coefficient for the model in different training stages, and (3) experimental results on three benchmark datasets with three widely used query strategies and two hyperparameter searching methods showing that CHAIN has a distinct and robust performance in most of the circumstances.

\section{Related Works}
\paragraph{Low-budget Active Learning.}
The goal of active learning (AL) is to render the best model given a limited budget of annotations. 
One of the fundamental components in AL is the query strategy, which selects a subset of instances from the pool of unlabeled data and queries their labels from an oracle~\citep{settles2009active}. 
Existing query strategies can be categorized into uncertainty-based~\citep{uncertainty1, uncertainty2, uncertainty3}, representation-based~\citep{representation1, representation2, representation3} or hybrid~\citep{hybrid1, hybrid2, hybrid3}. 
However, in the low-budget regime (e.g., less than 400 instances are labeled), the sub-optimality of the heuristic can lead to significant loss~\citep{mahmood2021low}, which is called \textit{cold-start} phenomenon.
To tackle the \textit{cold-start} issue, \citet{mahmood2021low} proposed a core set query strategy which minimized the discrete Wasserstein distance between the selected and unlabeled, while~\citet{hacohen2022opposite} proposed a clustering algorithm for querying most typical instances from the max density region with diversity. 
Both of these methods partially alleviated the \textit{cold-start} phenomenon by proposing new query strategies but overlooked the small-sample bias issue brought by using MLE in model training. In this work, instead of developing a new query strategy to compete with existing ones, we focus on the small-sample bias during model training and propose a general method that could improve over a given query strategy in low-budget AL.

\paragraph{Firth Bias Reduction.}
Firth bias reduction focuses on reducing the bias caused by maximum likelihood estimators while the dataset is small, known as Firths' Penalized Maximum Likelihood Estimator (PMLE)~\citep{firth1993bias}. In the case of the exponential family of distributions, it has a simplified form that penalizes the likelihood by Jeffrey's invariant prior~\citep{firth1993bias, poirier1994jeffreys}, which is proportional to the determinant of the Fisher Information Matrix $F$. For logistic and cosine classifiers~\citep{chen2019closer}, which are widely used in few-shot classification, since they belong to the exponential family, Firth bias reduction can be further cast as adding a log-determinant penalty ($\log(\det(F))$) to the cross-entropy loss.
In this work, we, for the first time, explore the utility of Firth bias reduction in reducing the small-sample bias in low-budget AL.

\section{Preliminary}
Assume a $K$-way classification task and let $\mathcal{U}=\{\textbf{x}_1,...,\textbf{x}_N\}$ be the input unlabeled dataset with unknown ground truth $Y=[y_1,...,y_N]$. Given a limited budget of annotations, the goal of active learning (AL) is to wisely spend the budget on querying annotations from an \textit{oracle} (e.g., human expert) for model training. Specifically, we have $R>1$ rounds of querying, and each time we query annotations for $M$ instances according to some query strategy.

One of the popular query strategies is the uncertainty-based query, wherein different measurements of the model's uncertainty (e.g., entropy, the least confidence) on each instance can be leveraged.
Given a specific query strategy $Q$, at round $r$ we select the top-$M$ uncertain instances from the remaining unlabeled instances. 
We include them in training set after obtaining their labels from the oracle~\citep{uncertainty1, uncertainty2, uncertainty3}. The resultant training set $\mathcal{D}_r$ in round $r$ is in turn used to train a classifier $f_{r}(\cdot;\bm\beta_r)$ parameterized by $\bm\beta_r$. We would omit the subscript $r$ when there is no confusion.
In this work, we focus on low-budget AL, where the budget is very low compared to the size of the pool of unlabeled data.
In such a regime, each time model is trained with only a small set of labeled data, and usually with the Maximum Likelihood Estimation (MLE) (\ie, the cross-entropy loss).
As a consequence, the optimized model parameter $\bm\beta$ would be biased~\citep{firth1993bias} and hurt both the model performance and the quality of the queries in the latter rounds, especially those uncertainty-based query strategies that largely rely on the trained model of last round to estimate the uncertainty of remaining unlabeled instances.

\section{Methods}
In this section, we describe the proposed method \Ours (\textbf{C}urriculum firt\textbf{H} bi\textbf{A}s reduct\textbf{I}o\textbf{N}) that aims to tackle the small-sample bias issue in low-budget active learning.
First, we show how to leverage the Firth bias reduction technique as a plug-in regularizer to reduce the small-sample bias in the model training procedure at each round of active learning.
Then, we propose an efficient approach to optimize the coefficient of the Firth regularizer without heavy parameter tuning.
Finally, we present a simple curriculum of the coefficient that adaptively adjusts the strength of the regularizer for the model at different training stages.

\subsection{Firth Bias Reduction}
Based on \citep{firth1993bias, ghaffari2021importance}, in this section, we introduce a penalized cross-entropy loss by adding the Firth regularizer to reduce the small-sample bias.
We denote the parameter of the logistic model as $\bm\beta$, and the assignment probability of the $i^{th}$ data to class $k$ can be represented as
\begin{equation}
    \textbf{P}_{i,k} = \text{Pr}(y_i=k|\textbf{x}_i; \bm\beta) = \frac{\exp(\beta_k^\top \textbf{x}_i)}{\sum_{k'=0}^K\exp(\beta_{k'}^\top \textbf{x}_i)}.
\end{equation}
The likelihood of the current training set $\mathcal{D}$ given the $\bm\beta$ is 
\begin{equation}
    \label{eq:pr}
    \text{Pr}(\textbf{y}|\mathcal{D};\bm\beta)=\prod_{i=1}^N\sum_{k=0}^K \mathds{1}[y_i=k]\cdot\textbf{P}_{i,k},
\end{equation}
where $\mathds{1}[\cdot]$ denotes the binary indicator function.
We use the maximum likelihood estimation (MLE) to acquire the optimal parameter $\hat{\bm\beta}_{\text{MLE}}$, which can be defined as
\begin{equation}
    \hat{\bm\beta}_{\text{MLE}}=\underset{\bm\beta}{\text{arg max}} \ \text{Pr}(\textbf{y}|\textbf{x};\bm\beta).
\end{equation}
Oftentimes the cross-entropy (CE) loss is used as the objective of the MLE, which can be interpreted as the log of Eq.~\ref{eq:pr}.
This draws out the optimization objective
\begin{align}
    \label{eq:opt_obj}
    \mathcal{L}_{\text{CE}}=-\sum_{i=1}^N l_{\text{CE}}(y_i\|\textbf{P}_{i,k})=-\sum_{i=1}^N\sum_{k=0}^K \mathds{1}[y_i=k]\log(\textbf{P}_{i,k}).
\end{align}
In logistic models, the penalized likelihood function proposed by Firth is equivalent to imposing Jeffery's prior~\citep{poirier1994jeffreys} on the parameters and making a maximum posterior estimation as
\begin{equation}
    \label{eq:pr_decom}
    \text{Pr}(\bm\beta|\mathcal{D},\textbf{y})=\frac{1}{Z}\cdot \text{Pr}(\textbf{y}|\mathcal{D};\bm\beta)\cdot \text{det}(F|r)^{\frac{1}{2}}
\end{equation}
where $Z$ is a normalization constant and $\text{det}(F|r)^{\frac{1}{2}}$ is the Jeffery's prior, the product of all $r$ non-zero eigenvalues of the Fisher information matrix $F$ to the power $1/2$, and $F$ is defined as the Hessian of the negative log-likelihood function with
\begin{equation}
    F=-\text{Hess}_{\bm\beta}(\mathcal{L}_{\text{CE}}) = \mathbb{E}_y[\nabla_{\bm\beta}\mathcal{L}_{\text{CE}}\cdot \nabla_{\bm\beta} \mathcal{L}_{\text{CE}}^\top].
\end{equation}
Taking the log of both sides of Eq.\ref{eq:pr_decom} yields the log-posterior as a sum of the cross-entropy loss and the Firth bias reduction term
\begin{equation}
    \mathcal{L}(\mathcal{D}; \bm\beta)=\log (\text{Pr}(\bm\beta|\mathcal{D},\textbf{y}))=\mathcal{L}_{\text{CE}}(\mathcal{D}; \bm\beta) + \lambda\cdot\mathcal{L}_{\text{Firth}}(\mathcal{D}; \bm\beta),
\end{equation}
where 
\begin{equation}
    \mathcal{L}_{\text{Firth}}=\frac{1}{2}\log (\text{det}(F|r)) + \text{const},
\end{equation}
and $\lambda$ is the coefficient of the regularizer.
Following~\citep{ghaffari2021importance}, we further acquire an easy-to-implement optimization objective for the Firth regularizer.
We firth decompose $F$ as
\begin{equation*}
    F_{(dK)\times (dK)}=X^\top_{(dK)\times(NK)}\cdot M_{(NK)\times(NK)}\cdot X_{(NK)\times(dK)},
\end{equation*}
where $d$ is the dimension of data feature space, $M$ is a block-diagonal matrix whose $i^{th}$ diagonal block is denoted as $M_i$.
By adopting the Determinant Amendment and Constant Dropping~\citep{ghaffari2021importance}, and using the Matrix-Determinant Lemma~\citep{harville1998matrix}, it comes out that 
\begin{align}
    \mathcal{L}_{\text{Firth}}&=\frac{1}{N}\sum_{i=1}^N\log(\text{det}(M_i)) + \text{Const} \notag\\
    &=\frac{1}{N}\sum_{i=1}^N\sum_{k=0}^K\log(\textbf{P}_{i,k}).
\end{align}

It is well known that Jeffery's prior encourages uniform class assignment probabilities~\citep{firth1993bias}.
Therefore, we treat the $\sum_{k=0}^K\log(\textbf{P}_{i,k})$ as a scaled average over a uniform distribution of the number of classes
\begin{equation}
    \mathcal{L}_{\text{Firth}}\propto\frac{1}{N}\sum_{i=1}^N\left[\sum_{k=0}^K \frac{1}{K+1}\cdot\log(\textbf{P}_{i,k})\right],
\end{equation}
and it further comes out that
\begin{equation}
    \mathcal{L}_{\text{Firth}} = \frac{-1}{N}\sum_{i=1}^N D_{\text{KL}}(\text{U}_{[0,K]}\|\textbf{P}_{i,k}) + \text{Const},
\end{equation}
where $\text{U}_{[0,K]}$ is the uniform distribution over the number of classes.
By combining with cross-entropy and dropping the constants, we have the final penalized optimization objective: 
\begin{equation}
    \label{eq:loss}
    \mathcal{L}=\frac{-1}{N}\sum_{i=1}^N\left[l_{\text{CE}}(y_i\|\textbf{P}_{i,k}) + \lambda \cdot D_{\text{KL}}(\text{U}_{[0,K]}\|\textbf{P}_{i,k})\right].
\end{equation}

\subsection{Bilevel Firth Coefficient Optimization}
According to Eq.~\ref{eq:loss}, we use $\lambda$ as a coefficient to adjust the strength of the Firth regularizer.
However, we found that the model performance is sensitive to the $\lambda$, because the variance of model performance with different $\lambda$ is around 4 points on average.
It indicates that one needs to perform some search algorithm to find the best $\lambda$. Furthermore, after each round of query, as new labeled instances are added and the size of the training set increases, we have to redo the search of the optimal $\lambda$, making the whole AL pipeline immensely time-consuming.

To attack this issue, we optimize $\lambda$ without heavy parameter tuning via bilevel optimization.
Specifically, let $\mathcal{D}$ and $\mathcal{V}$ be the current training set and a hold-out validation set, respectively.
The bilevel optimization objective can be formulated as

\begin{align}\label{eq:optobj}
    \lambda^* &= \argmin_{\lambda} \mathcal{L}_{\text{CE}}(\mathcal{V}; \bm\beta^*(\lambda)) \notag\\ 
    s.t. \ \bm\beta^*(\lambda)&=\argmin_{\bm\beta} \mathcal{L}_{\text{CE}}(\mathcal{D}; \bm\beta) + \lambda\cdot\mathcal{L}_{\text{Firth}}(\mathcal{D}; \bm\beta).
\end{align}
The above bilevel optimization problem is nontrivial to solve since computing the exact gradient of $\lambda$ regarding the objective of the outer-level optimization requires an optimal solution of inner optimization. 
Specifically, Eq.~\ref{eq:optobj} can be separated as the optimization of $\lambda^*$ for which the best response $\bm\beta^*(\lambda)$ is required, and the optimization of the best response $\bm\beta^*(\lambda)$.
For the optimization of $\bm\beta^*(\lambda)$, we adopt the SGD as an optimizer to update iteratively with time-step $t$
\begin{equation}
    \bm\beta^{(t+1)}:=\bm\beta^{(t)} - \eta\cdot\nabla_{\bm\beta^{(t)}}\Lt(\bm\beta^{(t)}, \lambda),
\end{equation}
where $\eta$ is the learning rate and $\Lt$ denotes the training loss.
Assume we have $T_2$ steps totally for update the $\bm\beta$, the best response$\bm\beta^*(\lambda)$ can be represented as
\begin{align}
    \label{eq:opt_beta}
    \bm\beta^*(\lambda) &\approx \bm\beta^{(T_2)} - \eta\cdot\nabla_{\bm\beta^{(T_2)}}\Lt(\bm\beta^{(T_2)}, \lambda) \\
    \text{where} \quad \bm\beta^{(T_2)} &:= \bm\beta^{(T_2-1)} - \eta\cdot\nabla_{\bm\beta^{(T_2-1)}}\Lt(\bm\beta^{(T_2-1)}, \lambda) \notag \\
    & \ \ \vdots \notag \\
    \text{where} \quad \bm\beta^{(1)} &:= \bm\beta^{(0)} - \eta\cdot\nabla_{\bm\beta^{(0)}}\Lt(\bm\beta^{(0)}, \lambda) \notag
\end{align}
To solve the optimization objective of $\lambda$, which requires the best response $\bm\beta^*(\lambda)$, we propose using an outer loop that wraps the optimization process for $\bm\beta^*(\lambda)$ described above.
We seek a value $\lambda^*$ which ensures that the training process $\Lt(\bm\beta, \lambda^*)$ produces parameter $\bm\beta^*(\lambda)$ which perform best against some metric $\Lv(\bm\beta^*(\lambda))$ we care about.
Essentially, an estimate of $\bm\beta^*(\lambda)$ is obtained following $T_2$ steps training as outlined in Eq.~\ref{eq:opt_beta}, which is then evaluated against $\Lv$.
The gradient of this evaluation, $\nabla_{\lambda}\Lv(\bm\beta^*(\lambda))$ is then obtained through backpropagation, and used to update $\lambda$.
Using $\tau=1,...,T_1$ as a time-step counter to symbolize the time-scale being different from that used in the \textit{inner-loop}, we formalize this update as
\begin{align}
    \label{eq:opt_lambda}
    \lambda^* &\approx \lambda^{(T_1)} - \gamma\cdot\nabla_{\lambda^{(T_1)}}\Lv(\bm\beta^*(\lambda^{(T_1)}) \\
    \text{where}\quad \lambda^{(T_1)} &:= \lambda^{(T_1-1)} - \gamma\cdot \nabla_{\lambda^{(T_1-1)}}\Lv(\bm\beta^*(\lambda^{(T_1-1)})) \notag \\
    & \ \ \vdots \notag \\
    \text{where}\quad \lambda^{(1)} &:= \lambda^{(0)} - \gamma\cdot \nabla_{\lambda^{(0)}}\Lv(\bm\beta^*(\lambda^{(0)})) \notag
\end{align}
where $\gamma$ is the learning rate, $\Lv$ denotes the validation loss, and $\bm\beta^*(\lambda)$ can be solved by Eq.~\ref{eq:opt_beta}. 
Implementing an iterative process as described in Eq.~\ref{eq:opt_lambda} will exploit the chain rule for partial derivatives in order to run back propagation. By decompose the $\nabla_{\lambda}\Lv(\bm\beta^*(\lambda))$, we have
\begin{equation}
    \nabla_{\lambda}\Lv(\bm\beta^*(\lambda)) = \nabla_{\bm\beta^{(t)}}\Lv(\bm\beta^*(\lambda)) \cdot \nabla_{\lambda}\bm\beta^*(\lambda)
\end{equation}
Obviously, $\nabla_{\bm\beta^{(t)}}\Lv(\bm\beta^*(\lambda))$ exists for $t\in\{0,...,T_2\}$. 
For $\nabla_{\lambda}\bm\beta^*(\lambda)$, the gradients trivially exist because: (1) $\lambda$ is a continuous hyperparameter and (2) Eq.~\ref{eq:opt_lambda} is a differentiable function of $\lambda$.
In this work, we use the GIMLI algorithm~\citep{grefenstette2019generalized} to implement the efficient calculation of $\nabla_{\lambda}\Lv(\bm\beta^*(\lambda))$.

\begin{algorithm}[t]
    \begin{small}
        \caption{\Ours}
    	\label{alg:chain}
        \KwIn{Unlabeled dataset $\mathcal{U}$; Query strategy $Q$; Size of each query $M$.}
        $\mathcal{D}_{1}\leftarrow$ Randomly sample $M$ instances from $\mathcal{U}$ and query the labels.\\
        $\mathcal{U}_{1}\leftarrow \mathcal{U} \setminus \mathcal{D}_{1}$\\
        \For{$r=1,2,...,R$}{
            \For{$t=1,2,...,T$}{
                \If{$t \ mod \ \Tout$ = 0}{
                    $\lambda_{t}^*\leftarrow$ solve Eq.~\eqref{eq:optobj}.
                }
                $\bm\beta^{(t+1)}\leftarrow $ perform a gradient descent step.
            }
            $\bm\beta^{*}_r\leftarrow \bm\beta^{(T)}$. \\
            $\mathcal{S}_{r+1}\leftarrow Q(\text{from}: \mathcal{U}_{r}, \text{size}: M, \text{model}: f(\cdot;\bm\beta^*_r))$. \\
            Query annotations of $\mathcal{S}_{r+1}$ from Oracle. \\
            $\mathcal{D}_{r+1}\leftarrow \mathcal{D}_{r} \cup \mathcal{S}_{r+1}$. \\
            $\mathcal{U}_{r+1}\leftarrow \mathcal{U}_{r} \setminus \mathcal{D}_{r+1}$. \\
        }
        \KwOut{Output final model $f(\cdot; \bm\beta^{*}_R)$.}
    \end{small}
\end{algorithm}

\subsection{The Curriculum of Firth Coefficient}
\begin{figure}
    \includegraphics[width=\linewidth]{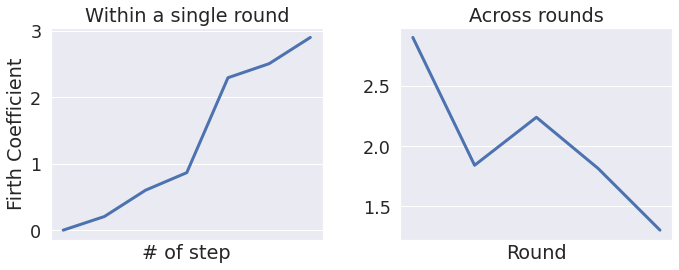}
    \caption{Example of the Firth coefficient curriculum. \textit{Left} is the coefficient curve in a single model training. \textit{Right} is the curve of the final coefficient of a model training across AL query rounds.}
    \label{fig:curriculum_exp}
\end{figure}

Regarding the impact of the Firth coefficient on the model performance, it is natural to hypothesize that the model at different training stages may require different $\lambda$ to obtain the best performance since the model may accumulate different levels of bias during the training. Therefore, we propose a simple curriculum of the Firth coefficient. That is, we repeatedly update the $\lambda$ via, again, solving the bilevel optimization but regarding the currently trained model.
Specifically, for every $T_2$ model training step, we use the current model parameters as the initialization of the $\bm\beta$ in the inner optimization of Eq.~\ref{eq:optobj} and then solve the bilevel optimization to get the new $\lambda$. In this way, the $\lambda$ is specific to the currently trained model and dynamically changes over the training process. 

As in Fig.~\ref{fig:curriculum_exp}, the coefficient we obtained by solving the bilevel optimization is increasing within a single model training. 
However, the final coefficient (the very last coefficient of a model training process) across different AL query rounds decreases with the increase in the size of the training set.
This phenomenon provides some insights into using a curriculum of $\lambda$: (1) within one model training, the bias is gradually increasing as the model saw a small set of training data multiple times, and therefore the $\lambda$ increases as well to strengthen the regularizer; (2) when we have more and more training data as the AL proceeds, the small-sample bias issue is naturally alleviated so the final $\lambda$ decreases. 
Finally, the overall algorithm of \Ours is summarized in Alg.~\ref{alg:chain}.

\begin{table*}[t!]
\centering
\caption{Final round performance comparison with and without \Ours of fine-tunning ResNet18 on CIFAR10, CIFAR100 and Fashion MNIST. \secbest{Blue} denotes the best result in each $M$, and \best{red} denotes the best results over all query strategies and $M$ in the corresponding dataset. Performance overall rounds are listed in the appendix.}
\label{tab:ft_final_round}
\begin{subtable}[h]{0.8\linewidth}
    \tiny
    \resizebox{\linewidth}{!}{%
    \begin{tabular}{l|l|llll|llll}
    \toprule\hline
    \multirow{2}{*}{Dataset($\downarrow$)}    & \multirow{2}{*}{Methods($\downarrow$)} & \multicolumn{4}{c|}{M=10}     & \multicolumn{4}{c}{M=20}      \\ \cline{3-10} 
                              &          & Entropy & CoreSet & BADGE & Random & Entropy & CoreSet & BADGE & Random \\ \hline
    \multirow{2}{*}{CIFAR10}  & Orig.  & \normal{52.26}   & \normal{52.84}   & \normal{53.04} & \normal{56.06}  & \normal{51.88}   & \normal{52.61}   & \normal{53.26} & \normal{54.43}  \\
                              & w/ CHAIN & 57.65   & 59.56   & 58.03 & \best{60.15}  & 55.90   & 56.90   & 56.94 & \secbest{57.47}  \\ \hline
    \multirow{2}{*}{CIFAR100} & Orig.  & \normal{15.52}   & \normal{14.63}   & \normal{16.50} & \normal{15.63}  & \normal{15.55}   & \normal{15.01}   & \normal{15.77} & \normal{15.79}  \\
                              & w/ CHAIN & 17.13   & 16.12   & 17.82 & \secbest{18.08}  & 16.62   & 16.36   & 16.21 & \best{18.82}  \\ \hline
    \multirow{2}{*}{Fashion MNIST} & Orig.   & \normal{88.50} & \normal{89.44} & \normal{87.47} & \normal{90.08} & \normal{90.72} & \normal{89.93} & \normal{88.64} & \normal{90.04} \\
                              & w/ CHAIN & 91.08   & \secbest{93.41}   & 92.22 & 91.28  & 90.99   & 91.36   & \secbest{91.64} & 91.48  \\ \hline
    \end{tabular}%
    }
\end{subtable}
\begin{subtable}[h]{0.8\linewidth}
    \tiny
    \resizebox{\linewidth}{!}{%
    \begin{tabular}{l|l|llll|llll}
    \hline
    \multirow{2}{*}{Dataset($\downarrow$)}    & \multirow{2}{*}{Methods($\downarrow$)} & \multicolumn{4}{c|}{M=50}     & \multicolumn{4}{c}{M=100}     \\ \cline{3-10} 
                              &          & Entropy & CoreSet & BADGE & Random & Entropy & CoreSet & BADGE & Random \\ \hline
    \multirow{2}{*}{CIFAR10}  & Orig.  & \normal{53.57}   & \normal{53.51}   & \normal{54.23} & \normal{52.25}  & \normal{52.26}   & \normal{53.58}   & \normal{52.12} & \normal{54.01}  \\
                              & w/ CHAIN & 58.49   & 57.14   & 58.39 & \secbest{58.63}  & 57.48   & 57.37   & 58.56 & \secbest{58.74}  \\ \hline
    \multirow{2}{*}{CIFAR100} & Orig.  & \normal{14.52}   & \normal{15.28}   & \normal{16.38} & \normal{16.63}  & \normal{14.49}   & \normal{15.68}   & \normal{16.76} & \normal{15.82}  \\
                              & w/ CHAIN & 15.95   & 17.25   & \secbest{17.58} & 16.83  & 14.69   & 16.05   & 17.51 & \secbest{17.70}  \\ \hline
    \multirow{2}{*}{Fashion MNIST} & Orig.  & \normal{89.20} & \normal{88.32} & \normal{91.88} & \normal{90.43} & \normal{90.32} & \normal{90.13} & \normal{90.11} & \normal{89.68} \\
                              & w/ CHAIN & 92.12   & 90.94   & \secbest{92.78} & 90.90  & 90.09   & 90.65   & \best{93.47} & 91.16  \\ \hline\bottomrule
    \end{tabular}%
    }
\end{subtable}
\end{table*}

\begin{figure*}[!ht]
    \centering
    \begin{subfigure}{\linewidth}
        \includegraphics[width=\linewidth]{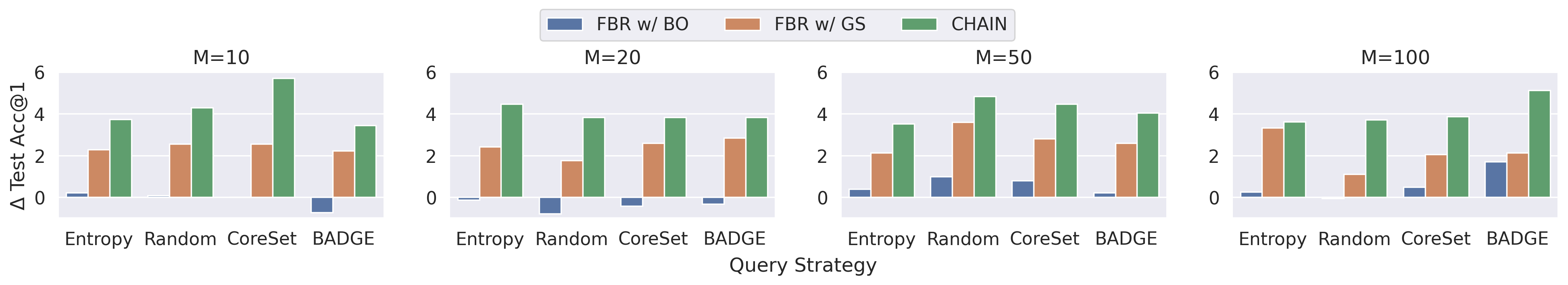}
        \caption{CIFAR10}
        \label{fig:cifar10_main}
    \end{subfigure}
    \begin{subfigure}{\linewidth}
        \includegraphics[width=\linewidth]{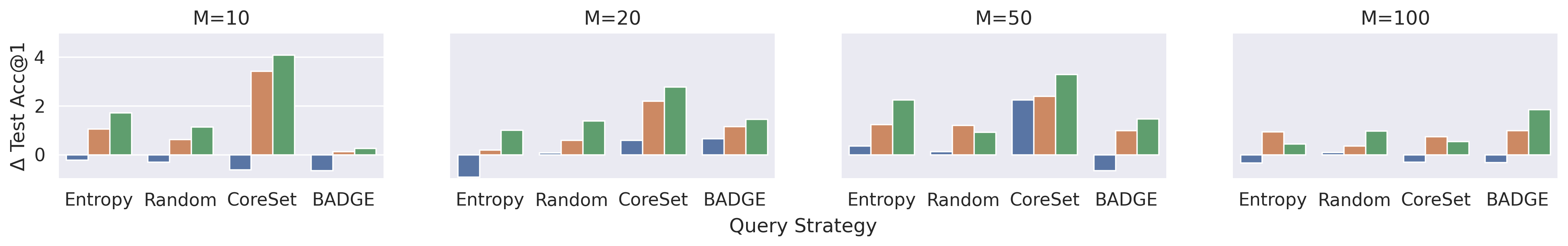}
        \caption{CIFAR100}
        \label{fig:cifar100main}
    \end{subfigure}
    \caption{Comparison of the performance gain over all rounds between FBR+BO, FBR+GS and CHAIN of fine-tunning ResNet18 on CIFAR10, CIFAR100. Both of the sub-figures share the same figure legend shown in Fig.\ref{fig:cifar10_main}. Results on Fashion MNIST are listed in the appendix.}
    \vspace{-1mm}
    \label{fig:main}
\end{figure*}

\section{Experiment}
In this section, we summarize the details of the experiment setting, including the datasets, AL setups, implementation details, and baselines.
We further provide the comparison result, ablation study, and Firth coefficient curriculum study on fine-tuning and logistic regression tasks.

\subsection{Experiment Setups}
\paragraph{Datasets and AL setups.}
We use three image classification benchmark datasets: CIFAR10~\citep{cifar10}, CIFAR100~\citep{cifar10} and Fashion MNIST~\citep{fashionmnist} to evaluate \Ours.
The datasets are separated into training and test sets originally, and we randomly sample a small-size validation set from the training set.
We study both deep learning and linear model settings.
For the former, we fix the model being trained to be ResNet-18~\citep{he2016deep} pre-trained on ImageNet~\citep{5206848}, while for the latter, we use the features extracted from also a ResNet-18 pre-trained on ImageNet and fit a logistic regression model on top of the extracted features. To demonstrate the efficacy of \Ours for reducing the small-sample bias in low-budget AL, we set our total budget as $500=M\times R$, where the budget of the training set at each query round is $M\in\{10, 20, 50, 100\}$, and total query round $R=500/M$.

\paragraph{Implementations.}
In the proposed \Ours, we have two major hyperparameters: $\Tin$ is the number of steps we used in solving the bilevel optimization, and within one model training, we solve the bilevel optimization to get a new $\lambda$ every $\Tout$ model training steps.
For the deep learning setting, we set $\Tin=50$ and $\Tout=100$ for all datasets.
For logistic regression, we set $\Tin=1$ and $\Tout=5$ for all the datasets. We use SGD with a momentum of $0.9$ for fine-tuning the ResNet-18 and Adam for training the logistic regression model.
We use $0.01$ as the learning rate for fine-tuning and $0.001$ for logistic regression.
For bilevel optimization, we use Adam as our optimizer, and we set the learning rate as $0.01$ for the deep learning setting and $0.05$ for the linear model setting.
To reduce the impact of changing the training set to the AL optimization process, we keep our batch size $|\mathcal{B}|$ positive proportional to the training set size $|\mathcal{S}_t|$, \ie,  $|\mathcal{B}|=0.1|\mathcal{S}_t|$. All our experiments are conducted on 6 Nvidia A6000 GPUs and Intel Xeon Gold 6226R CPU, and our code\footnote{We will release the source code after the paper is accepted.} is implemented in Python 3.8 and PyTorch 1.9.

\subsection{Baselines}
\Ours can be readily combined with any query strategy since it only modifies the model training procedure. 
To demonstrate the above claim, we choose three well-known uncertainty-based, representation-based, and hybrid query strategies and the naive random query strategy to show that \Ours can improve the model performance by reducing the small-sample bias in low-budget AL.
Besides those selected query strategies, we also compare \Ours with two widely used hyperparameter searching methods to show that \Ours performs better without efficiency loss.

\subsubsection{Query strategies}
\textbf{Entropy}~\citep{joshi2009multi}
Entropy is the most typical uncertainty-based query strategy.
Higher entropy values indicate greater uncertainty in the probability distribution.
\textbf{CoreSet}~\citep{sener2017active} is a representation-based query strategy that adopts the coreset selection based on data embeddings(representations).
\textbf{BADGE}~\citep{ash2019deep} is a hybrid query strategy incorporating both predictive uncertainty and data diversity into every selected batch.

\subsubsection{Hyperparameter searching methods}
To show the efficacy of our bilevel optimization approach for seeking the best Firth coefficient, we include the Firth bias reduction (FBR) with grid search over the coefficient as a baseline (marked as \textbf{FBR w/ GS}, GS: Grid Search).
To further verify the usefulness of the proposed curriculum of the coefficient, our last baseline is \textbf{FBR w/ BO} (BO: Bilevel Optimization), which only applies the bilevel optimization once before the model training process and uses the output $\lambda$ as a fixed hyperparameter during the model training.

\subsection{Main Result on Fine-tuning}
\label{sec:main_ft}
We evaluated \Ours against multiple baselines across four query strategies and three benchmark datasets, averaging the performance over three runs per query round. The results, summarized in Tab.~\ref{tab:ft_final_round}, showcase \Ours' performance enhancement for the chosen query strategies. We conducted a t-test to verify this improvement, taking $\mu_{\text{Orig.}}=\mu_{\text{w/ CHAIN}}$ as the null hypothesis and $\mu_{\text{Orig.}} < \mu_{\text{w/ CHAIN}}$ as the alternative. With $t$-values of $-15.52, -5.24$, and $-6.19$ for CIFAR10, CIFAR100, and Fashion MNIST, respectively, all values fall below the critical value $-1.6859$, strongly favoring the alternative hypothesis.
Notably, CIFAR10 and CIFAR100 perform best with a small $M$, likely due to their complex features and consequential small-sample bias. Conversely, Fashion MNIST excels with larger $M$, likely due to its simpler features, as highlighted in Tab.~\ref{tab:ft_final_round} (additional discussions in Sec.~\ref{sec:firth_curriculum} and the appendix).

We also compared the over-round performance gain of our method with other well-known hyperparameter searching methods across all datasets, and we recorded the results in Fig.~\ref{fig:main}.
Upon analyzing the results of \Ours, we observed that the performance per round was relatively consistent overall $M$ for entropy and random query strategies. However, for CoreSet and BADGE, the performance gain displayed a decreasing trend (for CoreSet) and an increasing trend (for BADGE), respectively.
This indicates that the CoreSet strategy is highly biased when the dataset is small, while the bias of BADGE increases as the dataset grows larger. 
Therefore, we recommend using BADGE with lower $M$ and CoreSet with higher $M$.





\begin{table*}[t]
\centering
\caption{Final round performance comparison with and without \Ours of logistic regression on CIFAR10. \secbest{blue} denotes the best result in each $M$, and \best{red} denotes the best results over all query strategies and $M$ in the corresponding dataset. Performance overall rounds are listed in the appendix.}
\label{tab:lr_final_round}
\begin{subtable}[t]{0.8\linewidth}
    \tiny
    \resizebox{\textwidth}{!}{%
    \begin{tabular}{l|l|llll|llll}
    \toprule\hline
    \multirow{2}{*}{Dataset($\downarrow$)} & \multirow{2}{*}{Methods($\downarrow$)} & \multicolumn{4}{c|}{M=10}     & \multicolumn{4}{c}{M=20}      \\ \cline{3-10} 
     &          & Entropy & CoreSet & BADGE & Random & Entropy & CoreSet & BADGE & Random \\ \hline
    \multirow{2}{*}{CIFAR10}   & Orig.  & \normal{28.28} & \normal{11.62} & \normal{43.03} & \normal{39.33} & \normal{31.27} & \normal{10.56} & \normal{40.18} & \normal{40.96} \\
     & w/ CHAIN & 37.47   & 32.15   & \secbest{44.12} & 43.44  & 37.79   & 33.90   & \secbest{44.86} & 41.36  \\ \hline
    \end{tabular}%
    }
\end{subtable}
\begin{subtable}[t]{0.8\linewidth}
    \tiny
    \resizebox{\textwidth}{!}{%
    \begin{tabular}{l|l|llll|llll}
    \hline
    \multirow{2}{*}{Dataset($\downarrow$)} & \multirow{2}{*}{Methods($\downarrow$)} & \multicolumn{4}{c|}{M=50}     & \multicolumn{4}{c}{M=100}     \\ \cline{3-10} 
     &          & Entropy & CoreSet & BADGE & Random & Entropy & CoreSet & BADGE & Random \\ \hline
    \multirow{2}{*}{CIFAR10}   & Orig.  & \normal{33.50} & \normal{11.49} & \normal{44.06} & \normal{40.35} & \normal{28.35} & \normal{11.51} & \normal{42.04} & \normal{39.87} \\
     & w/ CHAIN & 36.79   & 34.29   & \secbest{44.93} & 42.19  & 39.61   & 37.21   & \best{45.61} & 42.99  \\ \hline\bottomrule
    \end{tabular}%
    }
\end{subtable}
\end{table*}
\begin{figure*}[t]
    \begin{subfigure}{0.49\linewidth}
        \raggedleft
        \includegraphics[width=\linewidth]{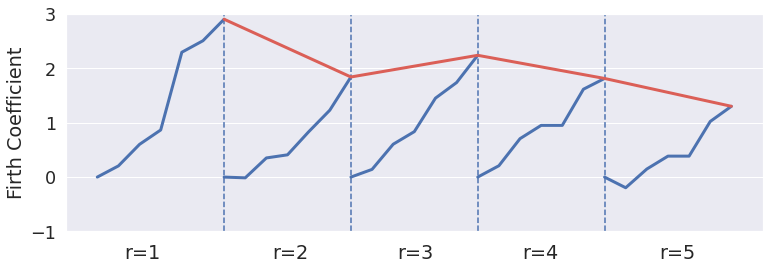}
        \caption{Fine-tuning a ResNet-18}
        \label{fig:lambda_changing_ft}
    \end{subfigure}
    \begin{subfigure}{0.49\linewidth}
        \raggedright
        \includegraphics[width=\linewidth]{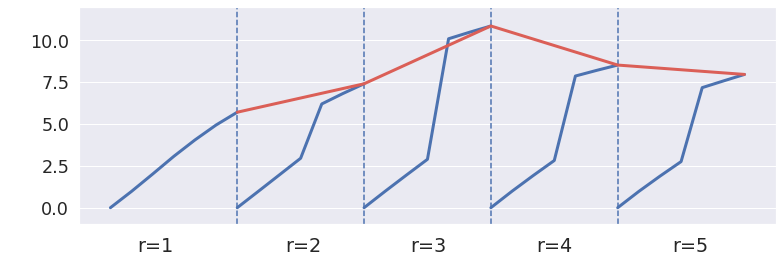}
        \caption{Logistic Regression}
        \label{fig:lambda_changing_lr}
    \end{subfigure}
    \caption{The trend of the Firth coefficient across model training steps and AL query rounds on CIFAR10 with random query strategy and $M=100$.}
    \label{fig:lambda_changing}
\end{figure*}
\begin{figure}[t]
    \centering
    \begin{subfigure}{0.49\linewidth}
        \centering
        \includegraphics[width=\linewidth]{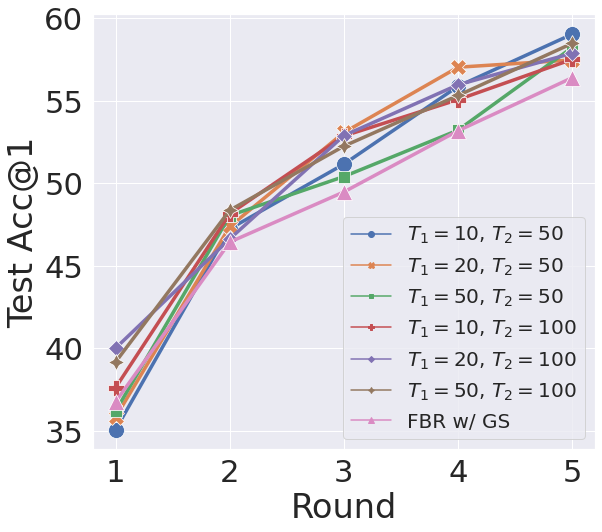}
        \label{fig:abla_cifar10}
        \caption{Fine-tuning a ResNet18}
    \end{subfigure}
    \begin{subfigure}{0.49\linewidth}
        \centering
        \includegraphics[width=\linewidth]{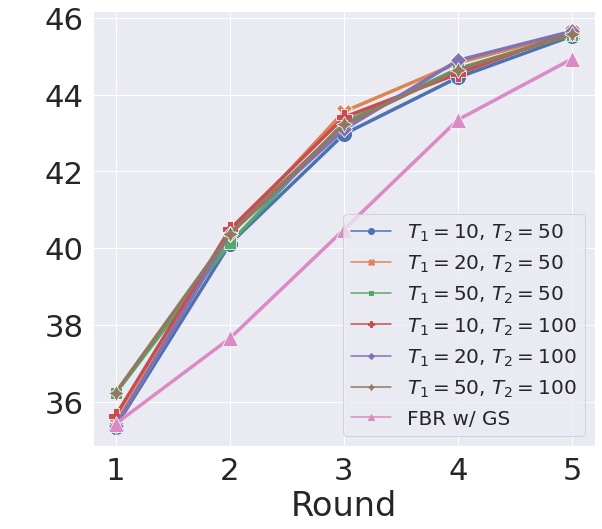}
        \label{fig:abla_fm}
        \caption{Logistic Regression}
    \end{subfigure}
    \caption{Ablation studies of the two hyperparameters $\Tin$ and $\Tout$ on CIFAR10 dataset with random query strategy and $M=100$. All the results are averaged over three runs.}
    \label{fig:abla}
\end{figure}

\subsection{Main Result on Logistic Regression}
To theoretically demonstrate the effectiveness of \Ours, we evaluated \Ours and other baseline methods using a logistic regression model on CIFAR10.
The results are presented in Tab.~\ref{tab:lr_final_round}, from which we can observe that \Ours consistently improves the overall performance across all query strategies.
Using the same t-test as in Sec.~\ref{sec:main_ft}, we obtained a $t$-value of $-3.32$, which is smaller than the critical value of $-t(N_{\text{Orig.}} + N_{\text{w/ CHAIN}}-2, 0.10)=-1.6859$. Based on this finding, we can conclude that the adoption of \Ours leads to a significant improvement in performance.
Upon further examining Tab.~\ref{tab:ft_final_round}, we observed that CoreSet failed to converge to good performance with logistic regression. 
However, a significant improvement was observed after adopting \Ours, indicating that the bias of CoreSet is amplified with less-parameterized models such as logistic regression.
On the other hand, BADGE has a strong bias-restricting ability, enabling it to achieve better performance even with less-parameterized models.
Further discussion of the performance of logistic regression can be found in the appendix.



\subsection{Ablation Study}
We provide ablation studies on two major hyperparameters: $T_1$ and $T_2$.
Tuning $T_1$ will influence the bilevel optimization, which determines the quality of the learned coefficient, while tuning the $T_2$ affects the frequency of the coefficient being updated during the model training.
We run CHAIN with varying $T_1\in \{10, 20, 50\}$ and $T_2\in\{50, 100\}$, and compare it with FBR w/ GS. 
The results can be found in Fig.~\ref{fig:abla}.
Following the results, CHAIN outperforms grid search in all combinations of $T_1$ and $T_2$.
We can see that performance of \Ours with different hyperparameters is more consistent with lower variance when using the logistic regression model than the deep model.
This observation indicates that the complex deep model is relatively more sensitive to the bilevel optimization process and the curriculum of the Firth coefficient.

\subsection{The Curriculum of the Firth Coefficient}
\label{sec:firth_curriculum}
We examine the progression of the optimization coefficient during model training and across various Active Learning (AL) rounds, with results depicted in Fig.~\ref{fig:lambda_changing} (details for CIFAR100 and Fashion MNIST are in the appendix). Both when fine-tuning deep models and training logistic regression models, the coefficient increases each round, eventually plateauing. This increase reflects the model's transition from a bias-free random initialization to a bias-impacted state due to small-sample size as training progresses, necessitating stronger regularization. The plateau likely arises when model parameters undergo minor changes, reflecting well-trained status, resulting in a stable coefficient.

Regarding the coefficient's dynamic across AL rounds, we observe a decrease in the final step coefficient during deep model fine-tuning. In contrast, logistic regression models see early-round coefficient increases followed by later-round decreases. This decrease in deep learning settings aligns with expectations: more labeled data naturally mitigate small-sample bias. The initial increase in the logistic regression model's coefficient is somewhat counterintuitive, as more data should typically decrease the need for Firth regularization. We posit that early query rounds trigger early stopping in fewer training steps for logistic regression models, resulting in lower final coefficients. In contrast, later rounds involve more training steps and subsequent coefficient updates, leading to higher coefficients.

\section{Conclusion}
In this work, we focus on the small-sample bias issue in low-budget active learning.
To mitigate the small-sample bias, we leverage the Firth bias reduction.
Since Firth bias reduction is sensitive to its coefficient, we adopt bilevel optimization to acquire the optimal coefficient for model training to avoid the time-consuming hyperparameter tuning.
Furthermore, we introduce a simple curriculum of the coefficient, which can adjust the regularization strength according to the current model training stage.
The experimental results show that CHAIN outperforms all baselines in most cases.
by comparing with three query strategies and two widely used hyperparameter searching methods on three datasets.

\bibliography{aaai22}

\clearpage

\section{Extra Experiment}

\begin{figure*}
\centering
\begin{subfigure}{\linewidth}
    \centering
    \includegraphics[width=\linewidth]{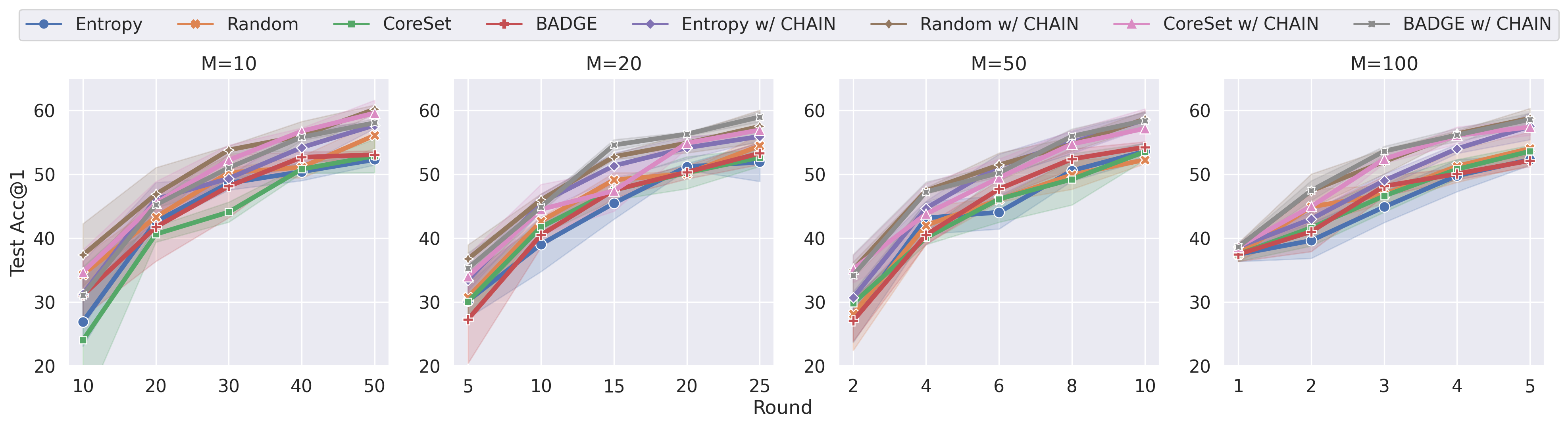}
    \caption{CIFAR10}
    \label{fig:cifar10_overround}
\end{subfigure}
\begin{subfigure}{\linewidth}
    \centering
    \includegraphics[width=\linewidth]{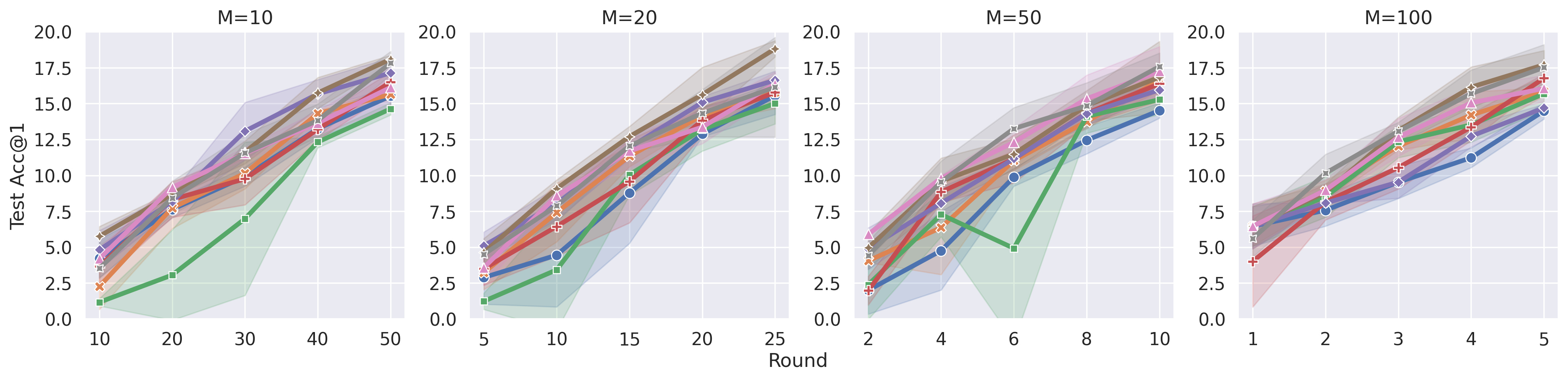}
    \caption{CIFAR100}
    \label{fig:cifar100_overround}
\end{subfigure}
\begin{subfigure}{\linewidth}
    \centering
    \includegraphics[width=\linewidth]{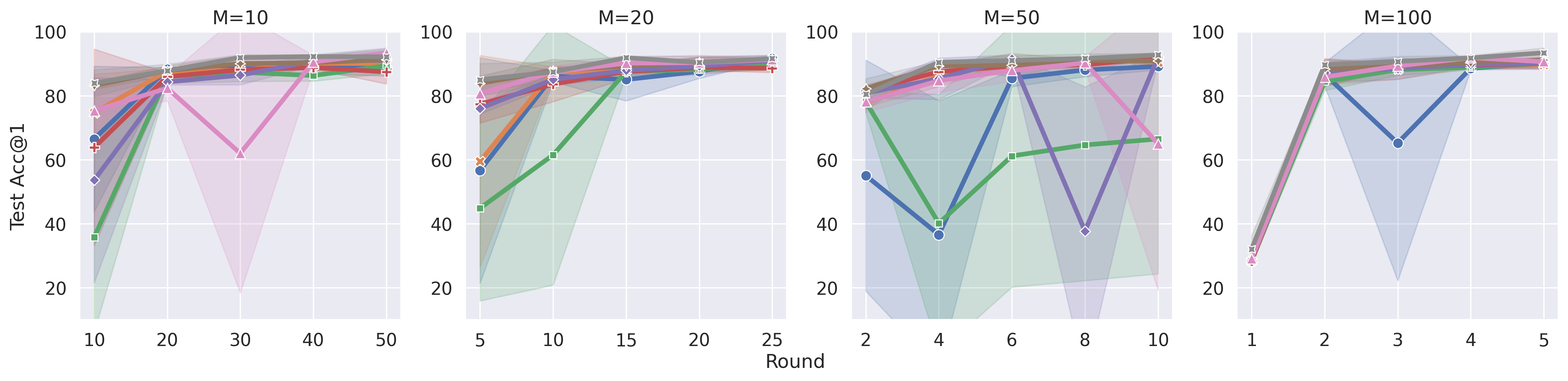}
    \caption{Fashion MNIST}
    \label{fig:fm_overround}
\end{subfigure}
\begin{subfigure}{\linewidth}
    \centering
    \includegraphics[width=\linewidth]{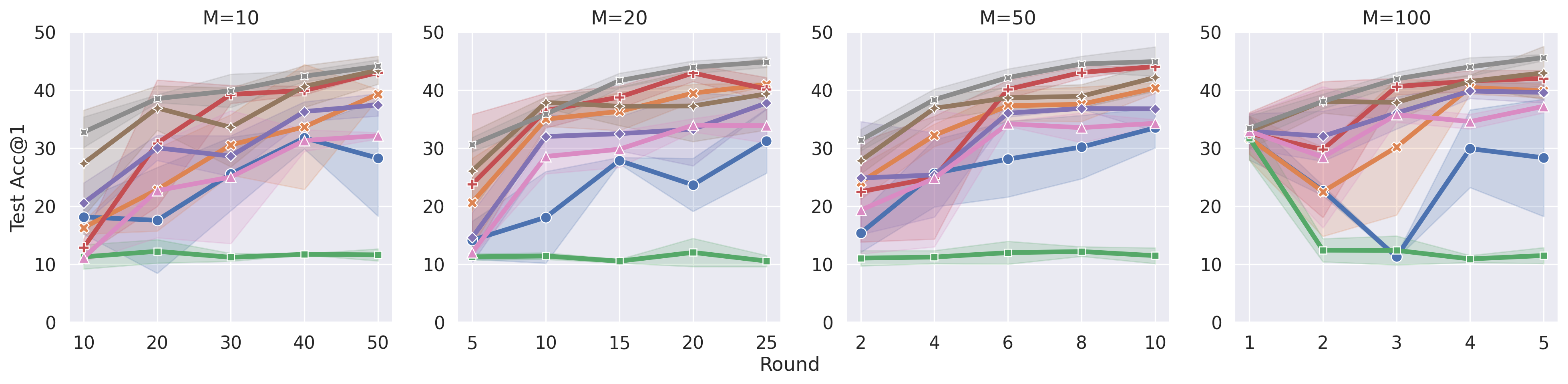}
    \caption{CIFAR10 (logistic regression)}
    \label{fig:cifar10_lr_overround}
\end{subfigure}
\caption{Per-round performance comparison (both of the fine-tuning and logistic regression) on three benchmark classification datasets: CIFAR10, CIFAR100, and Fashion MNIST. All of the above sub-figures share the same legend in Fig.~\ref{fig:cifar10_overround}. The solid lines are the average result over 3 runs, and the shadow area indicates the standard deviation.}
\label{fig:perround}
\end{figure*}

We provide the per-round comparison in Fig.~\ref{fig:perround}, and the final round results are reported in the main context.
To clarify the figure, we omit the curve of w/ GS and w/ BO of all query strategies and only report the final round results in Tab.\ref{tab:extra_main}.
We can see from the results that in most circumstances, training with a curriculum coefficient provides a promising performance gain over all rounds in CIFAR10 and CIFAR100, which is also reported in Fig.~\ref{fig:cifar10_main}.

\subsection{Other hyperparameter searching methods}
From Tab.~\ref{tab:extra_main}, we can find that FBR+GS also performs better than the origin under most situations but less than \Ours because of the limited searching space.
In this work, we search the $\lambda$ in \{0.0, 0.01, 0.1, 1.0, 3.0\} in each round, which means we will train five different models from scratch in each round, comparing their performance, and choose the best one to query data by the baseline query strategies from the dataset.
While FBR+GS yields good results, supporting such a search process requires huge performance, especially in active learning, which has multiple rounds, and we need to train a model from scratch at each round to prevent the influence of changing the distribution of the training set.
On the other hand, FBR+BO performs less than the original in many circumstances.
This is because bi-level optimization has a huge noise when the model is under-fitting, which is why we require $T_2$ steps in \Ours to get a "best response" of $\lambda$~\citep{lorraine2020optimizing}.

\subsection{Less perform in Fashion MNIST}
\begin{figure*}[!t]
    \includegraphics[width=\linewidth]{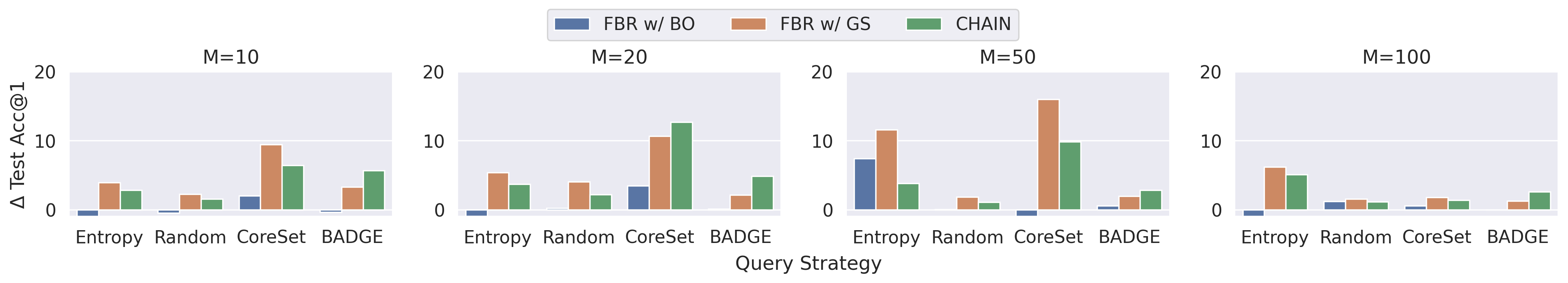}
    \caption{Comparison of the performance gain over all rounds between FBR+BO, FBR+GS and CHAIN of fine-tunning ResNet18 on Fashion MNIST.}
    \label{fig:fm_all}
\end{figure*}

\begin{figure*}[!t]
    \begin{subfigure}{0.49\linewidth}
        \raggedleft
        \includegraphics[width=\linewidth]{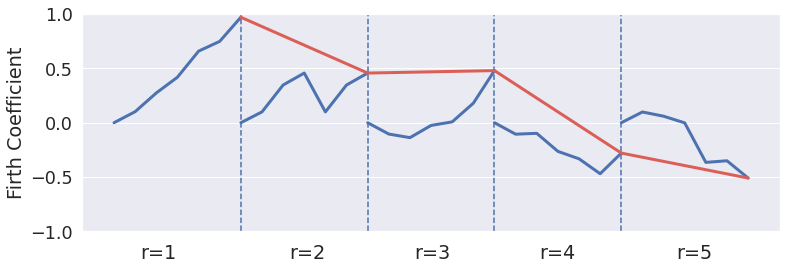}
        \caption{Fine-tuning a ResNet-18 on CIFAR100}
        \label{fig:lambda_changing_cifar100_ft}
    \end{subfigure}
    \begin{subfigure}{0.49\linewidth}
        \raggedleft
        \includegraphics[width=\linewidth]{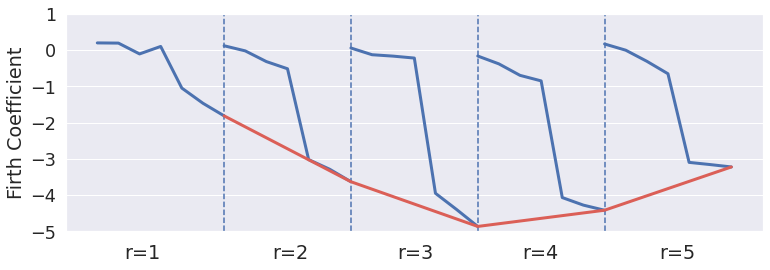}
        \caption{Fine-tuning a ResNet-18 on Fashion MNIST}
        \label{fig:lambda_changing_fm_ft}
    \end{subfigure}
    \caption{The trend of the Firth coefficient across model training steps and AL query rounds on CIFAR100 and Fashion MNIST with random query strategy and $M=100$. }
    \label{fig:lambda_changing_cifar100}
\end{figure*}

In the Fashion MNIST, the over-round performance gain of \Ours is sometimes less than FBR w/ GS, but \Ours still achieves the best final-round result in most circumstances.
By deep into the curriculum of the $\lambda$ of \Ours when performing on Fashion MNIST (Fig.~\ref{fig:fm_all}), we observe that \Ours tends to learn a negative value of $\lambda$, which according to Eq.~\ref{eq:loss} means that the model tends to "sharpen" the output confidence by minimizing the KL-divergence between the uniform distribution and the output confidence.
This observation indicates that the model can capture the features of Fashion MNIST under a low-budget regime with only a few steps, such that the model's performance on the training set can be improved by giving a more confident output (reflected in a sharp confidence distribution). 
However, such an optimization process will speed up the over-fitting, leading to less performance of \Ours at the start of training.
Fortunately, the "over-fitting" process can be mitigated by \Ours in the later round, which can be observed in Fig.~\ref{fig:lambda_changing_fm_ft} later round.
Moreover, this can be further alleviated by increasing the query rounds so that the \Ours achieve the best average performance in $M=10$ on Fashion MNIST.

\subsection{CoreSet versus BADGE in logistic regression}
We found that in logistic regression, the CoreSet cannot converge in CIFAR10 overall situations, and the bad performance appears consistently.
This is because CoreSet is a representation-based query strategy, and in the logistic regression, we froze the parameter of the feature extractor (ResNet18), which means the representation is fixed during training.
As a result, CoreSet will always query data with noised embeddings, leading to a bad performance.
While in the first round, the data is queried randomly to eliminate the bias of dataset initialization, so the coreset achieves better performance in the first round, and the performance gain is eliminated by the later selected noise data points.
However, BADGE does not suffer from the invariant embeddings even though it leverages the same embeddings of CoreSet.
This is because BADGE encodes the gradient, a representation of uncertainty, of the current model (logistic regression) into the embeddings, which helps the model query the data with large uncertainty, which can provide the most information gain.

\begin{table*}[h]
\caption{Final round performance comparison with and without \Ours of fine-tunning ResNet18 on CIFAR10, CIFAR100 and Fashion MNIST. \secbest{Blue} denotes the best result in each $M$, and \best{red} denotes the best results over all query strategies and $M$ in the corresponding dataset.}
\label{tab:extra_main}
\centering
\begin{subtable}[h]{0.9\linewidth}
    \tiny
    \resizebox{\textwidth}{!}{%
    \begin{tabular}{l|l|llll|llll}
    \toprule\hline
    \multirow{2}{*}{Dataset($\downarrow$)}    & \multirow{2}{*}{Methods($\downarrow$)} & \multicolumn{4}{c|}{M=10}     & \multicolumn{4}{c}{M=20}      \\ \cline{3-10} 
                              &           & Entropy & CoreSet & BADGE & Random & Entropy & CoreSet & BADGE & Random \\ \hline
    \multirow{4}{*}{CIFAR10}  & Orig.     & \normal{52.26}   & \normal{52.84}   & \normal{53.04} & \normal{56.06}  & \normal{51.88}   & \normal{52.61}   & \normal{53.26} & \normal{54.43}  \\
                              & w/ FBR+BO & 51.02   & 51.98   & 54.26 & 55.25  & 54.32   & 50.92   & 56.98 & 54.86  \\
                              & w/ FBR+GS & 53.68   & 54.31   & 56.90 & 57.63  & 56.00   & 56.22   & 56.37 & 56.42  \\
                              & w/ CHAIN  & 57.65   & 59.56   & 58.03 & \best{60.15}  & 55.90   & 56.90   & 56.94 & \secbest{57.47}  \\ \hline
    \multirow{4}{*}{CIFAR100} & Orig.     & \normal{15.52}   & \normal{14.63}   & \normal{16.50} & \normal{15.63}  & \normal{15.55}   & \normal{15.01}   & \normal{15.77} & \normal{15.79}  \\
                              & w/ FBR+BO & 14.77   & 14.32   & 15.92 & 16.73  & 14.39   & 15.44   & 16.21 & 16.20  \\
                              & w/ FBR+GS & 16.01   & 15.59   & 15.65 & 16.17  & 15.34   & 15.94   & 16.22 & 16.17  \\
                              & w/ CHAIN  & 17.13   & 16.12   & 17.82 & \secbest{18.08}  & 16.62   & 16.36   & 16.21 & \best{18.82}  \\ \hline
    \multirow{4}{*}{Fashion MNIST} & Orig. & \normal{88.50} & \normal{89.44} & \normal{87.47} & \normal{90.08} & \normal{90.72} & \normal{89.93} & \normal{88.64} & \normal{90.04} \\
                              & w/ FBR+BO & 88.66   & 86.99   & 90.30 & 91.81  & 88.63   & 89.46   & 90.15 & 90.45  \\
                              & w/ FBR+GS & 89.72   & 90.54   & 91.16 & 89.46  & 90.44   & 90.85   & 91.50 & 90.06  \\
                              & w/ CHAIN  & 91.08   & \secbest{93.41}   & 92.22 & 91.28  & 90.99   & 91.36   & \secbest{91.64} & 91.48  \\ \hline
    \end{tabular}%
    }
\end{subtable}
\begin{subtable}[h]{0.9\linewidth}
    \tiny
    \resizebox{\textwidth}{!}{%
    \begin{tabular}{l|l|llll|llll}
    \hline
    \multirow{2}{*}{Dataset($\downarrow$)}    & \multirow{2}{*}{Methods($\downarrow$)} & \multicolumn{4}{c|}{M=50}     & \multicolumn{4}{c}{M=100}     \\ \cline{3-10} 
                              &           & Entropy & CoreSet & BADGE & Random & Entropy & CoreSet & BADGE & Random \\ \hline
    \multirow{4}{*}{CIFAR10}  & Orig.     & \normal{53.57}   & \normal{53.51}   & \normal{54.23} & \normal{52.25}  & \normal{52.26}   & \normal{53.58}   & \normal{52.12} & \normal{54.01}  \\
                              & w/ FBR+BO & 53.55   & 53.77   & 54.20 & 54.88  & 51.70   & 54.65   & 54.85 & 54.39  \\
                              & w/ FBR+GS & 54.89   & 56.65   & 56.59 & 55.24  & 54.63   & 56.77   & 56.36 & 55.89  \\
                              & w/ CHAIN  & 58.49   & 57.14   & 58.39 & \best{58.63}  & 57.48   & 57.37   & 58.56 & \best{58.74}  \\ \hline
    \multirow{4}{*}{CIFAR100} & Orig.     & \normal{14.52}   & \normal{15.28}   & \normal{16.38} & \normal{16.63}  & \normal{14.49}   & \normal{15.68}   & \normal{16.76} & \normal{15.82}  \\
                              & w/ FBR+BO & 13.72   & 16.13   & 16.58 & 16.46  & 14.04   & 15.93   & 14.92 & 16.36  \\
                              & w/ FBR+GS & 15.37   & 16.33   & 16.58 & 16.57  & 14.22   & 16.39   & 16.36 & 16.51  \\
                              & w/ CHAIN  & 15.95   & 17.25   & \secbest{17.58} & 16.83  & 14.69   & 16.05   & 17.51 & \secbest{17.70}  \\ \hline
    \multirow{4}{*}{Fashion MNIST} & Orig. & \normal{89.20} & \normal{88.32} & \normal{91.88} & \normal{90.43} & \normal{90.32} & \normal{90.13} & \normal{90.11} & \normal{89.68} \\
                              & w/ FBR+BO & 90.32   & 89.59   & 89.87 & 90.76  & 89.77   & 90.55   & 89.39 & 90.89  \\
                              & w/ FBR+GS & 90.44   & 91.25   & 91.14 & 91.56  & 90.11   & 91.73   & 90.66 & 90.35  \\
                              & w/ CHAIN  & 92.12   & 90.94   & \secbest{92.78} & 90.90  & 90.09   & 90.65   & \best{93.47} & 91.16  \\ \hline\bottomrule
    \end{tabular}%
    }
\end{subtable}
\end{table*}

\end{document}